\documentclass[10pt, a4paper]{article}
\usepackage{lrec}
\usepackage{graphicx}
\usepackage{tabularx}
\usepackage{soul}
\usepackage{epstopdf}
\usepackage[utf8]{inputenc}

\usepackage{hyperref}
\usepackage{xstring}
\usepackage{amsmath}
\usepackage{color}
\usepackage{dirtytalk}

\usepackage{natbib}
\setcitestyle{numbers,square}

\newcommand{\newcite}{\cite}
\newcommand{\citelanguageresource}{\cite}

\title{A Tool for Facilitating OCR Postediting in Historical Documents}

\name{Alberto Poncelas, Mohammad Aboomar, Jan Buts, James Hadley, Andy Way}

\address{ADAPT Centre, School of Computing, Dublin City University, Ireland\\
        Trinity Centre for Literary and Cultural Translation, Trinity College Dublin, Ireland \\
         \{alberto.poncelas, jan.buts andy.way\}@adaptcentre.ie\\
          \{ABOOMARM, HADLEYJ\}@tcd.ie\\ }

\abstract{
Optical character recognition (OCR) for historical documents is a complex procedure subject to a unique set of material issues, including inconsistencies in typefaces and low quality scanning. Consequently, even the most sophisticated OCR engines produce errors. This paper reports on a tool built for postediting the output of Tesseract, more specifically for correcting common errors in digitized historical documents. The proposed tool suggests alternatives for word forms not found in a specified vocabulary. The assumed error is replaced by a presumably correct alternative in the post-edition based on the scores of a Language Model (LM). The tool is tested on a chapter of the book \textit{An Essay Towards Regulating the Trade and Employing the Poor of this Kingdom} \citelanguageresource{cary1719essay}. As demonstrated below, the tool is successful in correcting a number of common errors. If sometimes unreliable, it is also transparent and subject to human intervention.  \\ \newline 
}

\begin{document}

\maketitleabstract

\section{Introduction}

Historical documents are conventionally preserved in physical libraries, and increasingly made available through digital databases. This transition, however, usually involves storing the information concerned as images. In order to correctly process the data contained in these images, they need to be converted into machine-readable characters. This process is known as optical character recognition (OCR). Converting a book from image into text has obvious benefits regarding the identification, storage and retrieval of information. However, applying OCR usually generates noise, misspelled words and wrongly recognised characters. It is therefore often necessary to manually postedit the text after it has undergone the automatic OCR process. Usually, the errors introduced by the OCR tool increase with the age of the document itself, as older documents tend to be in worse physical condition. The circumstances of digitization, e.g. the quality of the scan and the mechanical typeset used, also impact the outcome of the OCR procedure. This paper proposes a tool for automatically correcting the majority of errors generated by an OCR tool. String-based similarities are used to find alternative words for perceived errors, and a Language Model (LM) is used to evaluate sentences. This tool has been made publicly available.\footnote{\url{https://github.com/alberto-poncelas/tesseract_postprocess}}

The performance of the tool is evaluated by correcting the text generated when using OCR with the book \textit{An Essay Towards Regulating the Trade and Employing the Poor of this Kingdom} \citelanguageresource{cary1719essay}.

\section{Related Work}

To improve the outcome of OCR, one can either focus on the processing of images in the scanned book, or on editing the output of the OCR tool. For either stage, several approaches have been proposed.

The approaches involving image-processing perform modifications on the scanned book that make the OCR perform better. Examples of these approaches include adding noise, as through rotation, for augmenting the training set \cite{bieniecki2007image}, reconstructing the image of documents in poor condition \cite{maekawa2019improving}, clustering similar words so they are processed together \cite{kluzner2009word} or jointly modeling the text of the document and the process of rendering glyphs \cite{berg2013unsupervised}.

Techniques for increasing accuracy by performing post-OCR corrections can be divided into three sub-groups. The first group involves lexical error correction, and consists of spell-checking the OCR output using dictionaries, online spell-checking \cite{bassil2012ocr}, and using rule-based systems for correcting noise \cite{thompson2015customised}. The second group of strategies for correcting OCR output is context-based error correction, in which the goal is to evaluate the likelihood that a sentence has been produced by a native speaker by using an {\em n}-gram LM to evaluate the texts produced by the OCR \cite{zhuang2004chinese}, and to use a noisy-channel model \cite{brill2000improved}, or a Statistical Machine Translation engine \cite{afli2016using} to correct the output of the OCR. A final approach proposes using several OCR tools and retrieving the text that is most accurate \cite{volk2010reducing,schafer2012combining}.

\section{OCR Challenges for Historical Document}
\label{sec:challenges}

Performing OCR is a challenging task. Although ideally the procedure should successfully generate the text represented in an image, in practice the tools often produce errors \cite{lopresti2009optical}. In addition, when older documents are converted into text further difficulties arise that cause the performance of the OCR tools to decrease. One of the problems of historical documents is that the quality of the print medium has often degraded over time. The quality of the paper also impacts the output, as in some cases the letters on the reverse side of a page are visible in the scanned image, which adds noise to the document.

\begin{figure}[h]

\includegraphics[width=8.5cm]{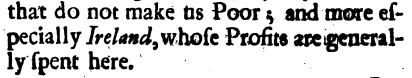}
\caption{Example of the scan of the book \textit{An Essay Towards  Regulating the Trade}. }
\label{fig:text_example}
\end{figure}

Furthermore, OCR systems are generally best suited to contemporary texts, and not built to handle the typefaces and linguistic conventions that characterize older documents. In Figure \ref{fig:text_example} we show a small extract of a scan of the book \textit{An Essay Towards  Regulating the Trade} to illustrate some of the problems frequently encountered. One may notice the following particularities of the text:

\begin{itemize}
    \item Some words such as \textit{especially}, \textit{whose} and \textit{spent} contain the \say{{\fontfamily{jkpvos}\selectfont{}s}} or long \say{s}, an archaic form of the letter \say{s} which can easily be confused with the conventional symbol for the letter \say{f}. 
    \item Some words, such as \textit{Poor} and \textit{Profits}, are capitalized even though they occur mid-sentence. This would be unusual in present-day English. 
\end{itemize}

\newcite{piotrowski2012natural} categorizes variations in spelling as uncertainty (digitization errors), variance (inconsistent spelling) and difference (spelling that differs from contemporary orthography). In our work we focus on the latter.

Spelling issues compound the general challenges touched upon before, such as the quality of the scan (e.g. the word \say{us} in Figure \ref{fig:text_example} is difficult to read even for humans). Further issues include the split at the end of the line (e.g. the word \say{especially} or \say{generally}).

\section{Proposed Tool}
\label{sec:proposal}

This paper introduces a tool that automatically edits the main errors in the output of an OCR engine, including those described in Section \ref{sec:challenges}. The method retains, next to the edited text, the material that has been automatically replaced. Thus, the human posteditor has the agency to approve or discard the changes introduced by the tool. The aim of automating the initial replacement procedure is to shorten the overall time spent on post-editing historical documents.

In order to execute the tool, run the command \texttt{ocr\_and\_postprocess.sh \$INPUT\_PDF  \$OUT \$INITPAGE \$ENDPAGE}. In this command, \texttt{\$INPUT\_PDF} contains the path of the \textit{pdf} file on which OCR will be performed, and \texttt{\$OUT} the file where the output will be written. \texttt{\$INITPAGE} and \texttt{\$ENDPAGE} indicate from which page until which page the OCR should be executed.

The output is a file consisting of two columns (tab-separated). The first column contains the text after OCR is applied and the errors have been corrected. In the second column, we include the list of edits performed by our tool, so that a human post-editor can easily identify which words have been replaced.

The pipeline of this approach is divided into three steps, as further explained in the subsections below. First, the OCR is executed (Section \ref{ref:step_ocr}). Subsequently, words that are unlikely to exist in English are identified and replacement words are sought (Section \ref{ref:step_word_altern}). Finally, the word-alternatives are evaluated within the larger sentence in order to select the best alternative (Section \ref{ref:step_word_repl}).

\subsection{Perform OCR}
\label{ref:step_ocr}

The first step is to extract part of the \textit{pdf} and convert it into a list of \textit{png} images (one image per page). These images are fed to an OCR engine and thus converted into text. The line-format in the text will conform to the shape of the image, meaning that word forms at the end of a line ending on an \say{-} symbol need to be joined to their complementary part on the following line to ensure completeness.

\subsection{Get alternative words}
\label{ref:step_word_altern}

As the output of the OCR tool is expected to contain errors, this text is compared to a list of English vocabulary referred to as \textit{recognized words}.

Once the text is tokenized and lowercased, some of the words can be replaced by alternatives that fit better within the context of the sentence. The words that we want to replace are those that are not included in the list of \textit{recognized words} or contain letters that are difficult to process by the OCR tool (as in the case of confusion between the letters \say{{\fontfamily{jkpvos}\selectfont{}s}} and \say{f} mentioned in Section \ref{sec:challenges}). For each of these words we construct a list of candidates for a potential replacement. This list is built as follows:

\begin{enumerate}

\item Even if a word seems to be included in the list \textit{recognized words}, it still may contain errors, as some letters are difficult for the OCR to recognize. As per the above, \say{f} can be replaced with \say{s}, and the resultant word can be added as a replacement candidate if it is a \textit{recognized word}.

\item If the word is not in the list \textit{recognized words}, we proceed along the following lines:
    \begin{enumerate}
    	\item The word is split into two subwords along each possible line of division. If both items resulting from the split are recognized words, the pair of words is added as an alternative candidate.
    	\item Similar words in the vocabulary are suggested using a string-distance metric. The 3 closest words, based on the \textit{get\_close\_matches} function of python's \textit{difflib} library, are included. 
    \end{enumerate}
    
\end{enumerate}

After this step, for a word $w_i$ we have a list of potential replacements $w_i, w_i^{(1)}, w_i^{(2)}... w_i^{(r_i)}$, where $r_i$ is the number of alternatives for $w_i$. Note also that the original word is included as an alternative.

\subsection{Replace words with their alternatives}
\label{ref:step_word_repl}

Once we have obtained a list of alternatives, we proceed to evaluate which of the alternatives fits best within the context of the sentence. This means that given a sentence consisting of a sequence of $N$ words $(w_1, w_2 ... w_N)$, the word $w_i$ is substituted in the sentence with each of its replacement candidates, and a set of sentences is obtained as in~\eqref{eq:sentence_list}:

\begin{equation} \label{eq:sentence_list}
\begin{split}
\{(w_1 \dotsc w_i \dotsc w_N), \\ (w_1 \dotsc w_i^{(1)} \dotsc w_N), \\ \dotsc \\ (w_1 \dotsc w_i^{(r_i)} \dotsc w_N) \}.
\end{split}
\end{equation}

The perplexity of an LM trained on an English corpus is used to evaluate the probability that a sentence has been produced by a native English speaker. Given a sentence consisting of a sequence of $N$ words as $w_1, w_2 ... w_N$, the perplexity of a language model is defined as in Equation \eqref{eq:perplexity}:

\begin{equation}\label{eq:perplexity}
PP=2^{-\frac{1}{n} P_{LM}(w_1...w_N) }
\end{equation}

Note that the LM evaluation is performed with lowercased sentences. Once the sentence with the lower perplexity has been selected, the case is reproduced, even if the word has been replaced. This is relevant in the case of capitalization conventions, as related to words such as \textit{Poor} or \textit{Profits} in Figure \ref{fig:text_example}.

\section{Experiments}
\label{sec:experiments}

\subsection{Experimental Settings}

In order to evaluate our proposal, we use the Tesseract\footnote{\url{https://github.com/tesseract-ocr/tesseract}} Tool \cite{smith2007overview} to apply OCR to the book \textit{An Essay Towards  Regulating the Trade}. Specifically, we convert into text a scan of the chapter \textit{An Act for Erecting of Hospitals and Work-Houses within the City of Bristol, for the better Employing and Maintaining the Poor thereof} (pages 125 to 139).

The list of \textit{recognized words} consists of the vocabulary of the python package nltk\footnote{\url{https://www.nltk.org/}}, expanded with a list of 467K words\footnote{\url{https://github.com/dwyl/english-words/blob/master/words.zip}} \citelanguageresource{dwyl}. For each word that is not included in the vocabulary list we search for the closest 3 alternatives (based on a string-distance metric).

In order to evaluate which word-alternative is the most plausible in the sentence we use a 5-gram LM built with KenLM toolkit \cite{heafield2011kenlm}, trained on the Europarl-v9 corpus \cite{koehn2005europarl}.

\subsection{Results}

The text obtained after applying OCR consists of 576 lines. These lines are usually short, containing about 7 words per line.

\begin{figure}[h]

\includegraphics[width=8.5cm]{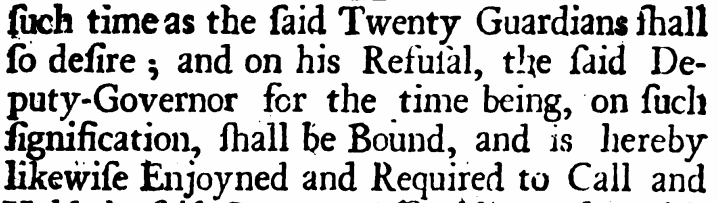}
\caption{Extract from the test set. }
\label{fig:example_Essay}
\end{figure}

\begin{table}[!htbp]
\centering
\begin{center}
\begin{tabular}{ |p{2.7cm}|p{2.7cm}|p{1.6cm}|}
\hline
Original	&    Edited	&    Changes    \\
\hline
\{uch timeas the faid Twenty Guardians fhall	&	\{uch times the said Twenty Guardians shall	&	timeas $\rightarrow$ times; faid $\rightarrow$ said; fhall $\rightarrow$ shall	\\
\hline
fo defire ; and on his Refutal, the faid	&	so desire; and on his Refutal, the said	&	fo $\rightarrow$ so; defire $\rightarrow$ desire; faid $\rightarrow$ said	\\
\hline
Deputy-Governor for the time being, on fuch	&	Ex-governor for the time being, on such	&	Deputy-Governor $\rightarrow$ Ex-governor; fuch $\rightarrow$ such	\\
\hline
fignification, fhall be Bound, and is hereby	&	fignification, shall be Bound, and is hereby	&	fhall $\rightarrow$ shall	\\
\hline
ikewife Enjoyned and Required to Call and	&	likewise Enjoined and Required to Call and	&	ikewife $\rightarrow$ likewise; Enjoyned $\rightarrow$ Enjoined	\\
\hline
\end{tabular}
\caption{ Example of postedited line }
\label{table:example_postedited}
\end{center}
\end{table}

In Figure \ref{fig:example_Essay} we show an extract of the scanned book. The text obtained after OCR is given in Table \ref{table:example_postedited} (in the first column).
Comparing the resultant text with the original, one can easily spot errors mentioned in Section \ref{sec:challenges},such as retrieving \say{fuch} instead of \say{such}, and further irregularities, such as interpreting \say{time as} as a single word.

Table \ref{table:example_postedited} also presents the text after being processed with our tool (second column). In the third column we include the substitution performed (this information is also retrieved by the tool). We observe that 66\% of the lines contain at least one correction. Each line has a minimum of 0 and a maximum of 3 corrections.

The tool is generally successful in correcting the words in which the letter \say{f} and \say{{\fontfamily{jkpvos}\selectfont{}s}} were previously confused. Most frequent in this regard are word-initial errors for \say{shall}, \say{so} and \say{said}, but word-internal mistakes, as in\say{desire} (see second row), are not uncommon.

In the first row we observe that the word \say{timeas} is not recognized as part of the vocabulary. The tool finds that the item can be split into the English words \say{time} and \say{as}. However, the tool also finds other options, and opts to render \say{times}, thus requiring human intervention and illustrating the necessity of transparency in the automated procedure. In the last row, a non-existent word has been corrected as \say{likewise}  because it is similar in terms of string-distance and is plausible according to the LM.

\begin{table}[!htbp]
\centering
\begin{center}
\begin{tabular}{ |p{1.5cm}|p{6cm}|}
\hline
Unrec. word	&   Alternatives \\
\hline
\say{faid}	&	\say{fai}, \say{f aid}, \say{fid}, \say{fa id}, \say{said}, \say{fraid}	\\
\hline
\say{timeas}	&	\say{timias}, \say{tim eas}, \say{time as}, \say{tineas}, \say{ti meas}, \say{times}	\\
\hline
\say{ikewife}	&	\say{likewise}, \say{ike wife}, \say{piewife}, \say{kalewife}	\\
\hline
\end{tabular}
\caption{ Example of replacement dictionaries }
\label{table:example_altern_dict}
\end{center}
\end{table}

Table \ref{table:example_altern_dict} presents some of the words that could not be found in the vocabulary (first column) and their respective candidates for replacement. The tool replaced these words by the most plausible alternative, employing th LM to evaluate the resulting sentence.

Despite numerous successful corrections, Table \ref{table:example_postedited} also shows some of the limitations of the tool. For example, the word \say{fignification} has not been properly replaced by a correct alternative. Other words have been incorrectly replaced, such as \say{Deputy-Governor}, which now occurs as \say{Ex-Governor}.

In our experiments, we observe that around 63\% of the errors are corrected by our tool. Most of the corrections are made in frequent words such as the word \say{shall} mentioned in Table \ref{table:example_postedited}.

\section{Conclusion and Future Work}

In this paper we have presented a tool to postprocess errors in the output of an OCR tool. As the problems addressed mainly pertains to historical documents, the tool was illustrated with reference to the early 18th-century text \textit{An Essay Towards  Regulating the Trade}. In order to achieve a more accurate representation of the original document than is commonly attained in image-text conversion, we constructed a system that identifies words that have potentially been incorrectly recognised and which suggests candidates for replacement. In order to select the best candidate, these alternatives are evaluated within the context of the sentence using an LM.

In this study we have manually stated which characters are misrecognized by the OCR system. In the future, we hope to develop a method for automatically identifying such characters.

We did not find large amounts of good-quality data from around 1700. Further research would benefit from LM models built on data from the same period as the test set, which could also be used to select appropriate sentences \cite{poncelas2016extending,poncelas2017applying}.

The tool could also be expanded to address related issues of textual organization, such as the automatic separation of side notes from a body of text. Overall, OCR technology is a fundamental factor in the dissemination of knowledge in the digital age, and to refine its output is essential.

\section{Acknowledgements}
The QuantiQual Project, generously funded by the Irish Research Council’s COALESCE scheme (COALESCE/2019/117).

This research has been supported by the ADAPT Centre for Digital Content Technology which is funded under the SFI Research Centres Programme (Grant 13/RC/2106).

% \nocite{*}
%\section{Bibliographical References}\label{reference}
%\label{main:ref}

%\bibliographystyle{lrec}
%\bibliographystyle{plain}
\bibliography{lrec2020W-xample-kc}

%\section{Language Resource References}
%\label{lr:ref}
%\bibliographystylelanguageresource{lrec}
%\bibliographylanguageresource{languageresource}
%\bibliography{languageresource}

\end{document}